%% file: bottleneck.tex
\documentclass[conference]{IEEEtran}
 \IEEEoverridecommandlockouts

\ifCLASSINFOpdf

\else

\fi

\input{includes}

\definecolor{colorAny}{HTML}{EEEEEE}
\definecolor{colorPRDGray}{HTML}{DDDDDD}
\definecolor{colorPRDarkBlue}{HTML}{0F4176}
\definecolor{colorPRMedBlue}{HTML}{147BBD}
\definecolor{colorPRLightBlue}{HTML}{66AED4}
\definecolor{colorPRGreen}{HTML}{97BF20}
\definecolor{colorPRLightGreen}{HTML}{BCD56F}

\usepackage{fancyhdr}
\renewcommand{\baselinestretch}{0.97} 

\newcommand\td{\tilde{d}}
\newcommand\tdmaxmax{\tilde{d}\subs{max,bound}}
\newcommand\tdmax{\tilde{d}\subs{max}}
\newcommand\dmax{{d}\subs{max}}

\begin{document}

\title{\uppercase{\bfseries Cooperative Automated Driving\\ for Bottleneck Scenarios\\ in Mixed Traffic}}

\author{M.V. Baumann$^{1\text{A}}$, J. Beyerer$^{1\text{B},2}$, H.S. Buck$^{3\text{, former }1\text{A}}$, B. Deml$^{1\text{C}}$, S. Ehrhardt$^{1\text{C}}$, Ch. Frese$^{2}$, D. Kleiser$^{2}$,\\M. Lauer$^{1\text{D}}$, M. Roschani$^{2}$, M. Ruf$^{4}$, Ch. Stiller$^{1\text{D}}$, P. Vortisch$^{1\text{A}}$, J.R. Ziehn$^{2}$\vspace{-9pt}%

\thanks{*This publication was written in the framework of the KAMO: Karlsruhe Mobility High Performance Center (\href{https://kamo.one}{kamo.one}), which is funded by the Ministry of Science, Research and the Arts and the Ministry of Economic Affairs, Labour and Housing in Baden-W\"urttemberg and as a national High Performance Center by the Fraunhofer-Gesellschaft.}%
\thanks{
\raggedright
$^{1}$Karlsruhe Institute of Technology (KIT), 76131 Karlsruhe, Germany
}%
\thanks{~~~~$^{\text{A}}$Institute for Transport Studies (IfV)}%
\thanks{~~~~$^{\text{B}}$Vision and Fusion Laboratory (IES)}%
\thanks{~~~~$^{\text{C}}$Institute for Human and Industrial Engineering (ifab)}%
\thanks{~~~~$^{\text{D}}$Department of Measurement and Control (MRT)}%
\thanks{
\raggedright
$^{2}$Fraunhofer IOSB, 76131 Karlsruhe, Germany,
{\tt\footnotesize jens.ziehn@iosb.fraunhofer.de}}%
\thanks{
\raggedright
$^{3}$platomo GmbH, 76137 Karlsruhe, Germany}%
\thanks{
\raggedright
$^{4}$Fraunhofer ICT, 76327 Pfinztal, Germany}%
}

\maketitle
\thispagestyle{fancy}

\begin{abstract}
Connected automated vehicles (CAV), which incorporate vehicle-to-vehicle (V2V) communication into their motion planning, are expected to provide a wide range of benefits for individual and overall traffic flow. A frequent constraint or required precondition is that compatible CAVs must already be available in traffic at high penetration rates. Achieving such penetration rates incrementally \emph{before} providing ample benefits for users presents a chicken-and-egg problem that is common in connected driving development. Based on the example of a cooperative driving function for bottleneck traffic flows (e.g. at a roadblock), we illustrate how such an evolutionary, incremental introduction can be achieved under transparent assumptions and objectives. To this end, we analyze the challenge from the perspectives of automation technology, traffic flow, human factors and market, and present a principle that {1)~accounts} for individual requirements from each domain; {2)~provides} benefits for any penetration rate of compatible CAVs between 0\,\% and 100\,\% as well as upward-compatibility for expected future developments in traffic; {3)~can} strictly limit the negative effects of cooperation for any participant and {4)~can} be implemented with close-to-market technology. We discuss the technical implementation as well as the effect on traffic flow over a wide parameter spectrum for human and technical aspects.
\end{abstract}

\section{Background and Motivation}

\begin{figure*}%
\includegraphics[width=\textwidth]{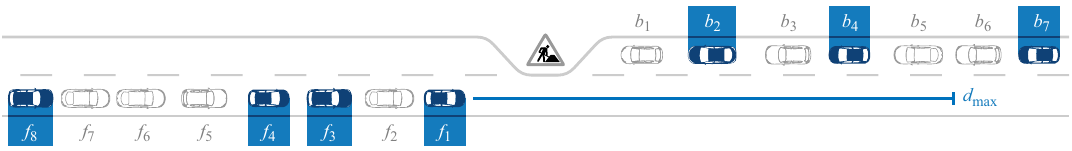}%
\caption{Example situation at the bottleneck, with connected automated vehicles (CAVs) shown as shaded, and regular (human-driven) vehicles shown as outlines. Vehicles on the blocked lane ($b_1, ..., b_7$) have to wait for a vehicle on the free lane ($f_1, ..., f_8$) to yield the right of way. In this paper, an optimal decision-making algorithm for the CAVs is sought, while human drivers are assumed to act stochastically. The algorithm requires each CAV on the free lane to pick a distance threshold $\dmax$, given here for the CAV $f_1$, specifying the maximum number of vehicles that the CAV would grant the right of way to.}%
\label{fig:example}%
\end{figure*}

Cooperative automated driving (CAD), i.e. the combination of vehicle-to-vehicle communication with the dynamic driving task, has been shown to provide a wide range of potentials, ranging from improved traffic flow, for example at intersections, to increased safety, for example in cooperative collision avoidance \cite{wang2019survey}. However, there remain challenges \cite{shladover2021opportunities}. In particular, most high-impact applications (e.g. \cite{Frese2011, frese2011comparison, geiger2012team, kamal2013coordination}) assume high penetrations of compatibly equipped Connected Automated Vehicles (CAV), and high levels of automation (SAE Level 4+). Reaching this state soon, and in an evolutionary way, is assumed to require CAD systems that satisfy the following conditions:
\begin{itemize}
	\item They must be able to operate at relatively low levels of automation and with little additional equipment;
	\item They must be able to provide \emph{immediate} added value (i.e. market value) for the user at relatively low penetrations of ``partner vehicles'' in mixed traffic (i.e. human drivers, non-connected automated vehicles and CAVs) and without extensive infrastructure requirements, and achieve sufficient acceptance for continuous use \cite{fank2017factors};
	\item They must continue to provide \emph{long-term} added value for the user over expected future developments; in particular their usefulness should not be \emph{limited} strictly to low penetrations of CAVs;
	\item They must contribute to establishing relevant, upward-compatible equipment in traffic that future generations of CAVs can build upon.
\end{itemize}

\subsection{Cooperative Bottleneck Resolution}

One such example is a CAD function that resolves bottleneck scenarios, for example at unsignalized construction sites, roadblocks, waste collection, accident sites or similar, when traffic in one direction has to move to the opposite lane, or when both directions overlap without priority in a narrow passage. These scenarios are especially difficult to manage for automated vehicles, as determining and obtaining the right of way often requires human interaction and, at high traffic volumes, a certain level of assertiveness.

We analyze how an appropriately designed CAD function, as presented in \cite{baumannbottleneck}, can resolve these scenarios effectively at low penetrations of compatible CAVs, establish a ``balanced'' flow over increasing levels of such vehicles, and provide adequate gains for users in the vehicles that \emph{receive} the right of way just as much as the users that \emph{grant} the right of way. The function is therefore studied from a technical perspective (including algorithms and V2X considerations), a traffic perspective (including the analysis of how the CAD function affects overall traffic flow and individual driving times), a human factors perspective (including appropriate human--machine interface (HMI) and acceptance) and a market perspective (including costs of operation and market value).

\subsection{Scope of the Paper}

The scope of the paper is thus to develop an algorithm that specifies the decision-making of CAVs at such a bottleneck situation, including a specification of what must be communicated via V2X, and how vehicles should react to this. The goal is to improve traffic quality, with respect to traffic flow rates (number of vehicles per driving direction per unit of time), ecology (reduced emissions by reduced accelerations and decelerations) and acceptance (e.g. achieving a balanced flow between driving directions, reducing individual waiting times or increasing predictability).

With regards to this, the scenario (described in detail in Sec.~\ref{sec:problem-definition}) differs notably by traffic volume: At low volumes, where vehicles only occasionally meet at the bottleneck, overall flow rates meet the demand well. However, individual delays can occur when two vehicles approach the bottleneck with similar timing. In this case, time may be lost to negotiate the right of way, and unnecessary accelerations and decelerations may occur. A solution for this case is presented in the previous paper \cite{naumann2017cooperativePlanning}, where two CAVs approaching a bottleneck perform minimal adjustments to their speeds to separate their arrival times and pass the bottleneck without notable delays or decelerations. At moderate traffic volumes (where sufficiently wide gaps in oncoming traffic occur regularly), a mixture of approaches (cf. Sec.~\ref{sec:conclusion}) can be required: Available gaps can be exploited (research concerning gap acceptance for human drivers in bottlenecks has been presented in \cite{ehrhardt2021gap}), but depending on their rate, a balanced traffic flow may still require vehicles on the free lane to occasionally yield the right of way. For our paper, we strictly focus on high traffic volumes where the right of way \emph{always} has to be yielded explicitly.

\section{Problem Definition and Requirement Specification}\label{sec:problem-definition}

For the two-way street shown in Fig.~\ref{fig:example}, we distinguish between vehicles on the \emph{blocked lane} which has to switch to the opposite lane due to the bottleneck, and vehicles on the \emph{free lane} which can drive straight through the bottleneck. No traffic lights are installed; it is assumed that by traffic laws, the free lane generally has the right of way, but can yield to the right of way to vehicles on the blocked lane.

We denote vehicles on the \emph{free} and on the \emph{blocked} lane by
\begin{equation}
F = (f_1, f_2, f_3, ...)\quad\text{and}\quad B = (b_1, b_2, b_3, ...)
\end{equation}
respectively. Since the system is designed purely as additional equipment to an automated vehicle, the vehicle is assumed to be equipped with basic technology that enables the automated operation in the given environment. The following technical prerequisites are taken to be available in the vehicle:
\begin{itemize}
	\item Systems for perception, planning and action to perform a regular (non-connected) dynamic driving task (DDT) automatically in the environment (SAE Level 1+). This will typically include detection of other traffic participants, drivable areas and lanes at sufficient range to enable the safe passage through the bottleneck in human traffic conditions. As by the system design (Sec.~\ref{sec:algorithm}), any such safety-critical feature will not be affected by the proposed system;
	\item A system for global or relative positioning on the road (GNSS and map data, SLAM, odometry, ...);
	\item Compatible cellular or short-range peer-to-peer communications (as in 802.11p, DSRC, WAVE, ETSI ITS-G5).
\end{itemize}

We also assume that each vehicle is aware of the existence, location and approximate size of the bottleneck. This can be provided naturally by other compatible CAVs which already passed the bottleneck, or by online map data. Its implementation is not part of the system design discussed here.

From the traffic perspective, the ideal state, under dense traffic conditions, is taken to be ``balanced flow'', where both sides of the bottleneck flow at the same rate.\footnote{Note that this is a simplifying assumption for the scope of this presentation. As stated in Sec.~\ref{sec:conclusion}, for notably different traffic \emph{demands} in each direction, a bias towards the higher-demand direction will actually usually be preferable. However, the same basic principles apply in both cases.} From this, the following requirements have been set for the system:

\begin{itemize}
	\item Traffic flow should approach balanced flow as penetrations of compatible CAVs increase;
	\item Negative emergent global effects in this process must be avoided; in particular, the system should not overcompensate by systematically disadvantaging connected vehicles or vehicles on the free lane, for any penetration up to 100\% of compatible CAVs (vs. other traffic participants);
	\item Negative individual effects for any traffic participant should be specified and limited;
	\item Acceptable added value should be provided for any active user of the system, both on the free lane and on the blocked lane;
	\item Additional technical requirements of the system should be negligible; in particular, the system should require no dedicated, centralized controller overseeing the traffic flow and no extensive computations;
	\item Safety-critical automated driving features should be unaffected by the system.
\end{itemize}

\section{Algorithm}\label{sec:algorithm}

Based on these requirements, the proposed algorithm is designed to leverage emergence, in the sense that simple local rules are specified that, on a larger scale, achieve the desired effects without global control or global optimization, and without the necessity to share complex information. To better motivate the actually proposed algorithm, we introduce several variants that do not satisfy the above requirements.

\subsection{The Non-Connected Variant}\label{sec:var-nonconnected}

This variant attempts to satisfy the requirements without use of V2X communications, hence the ``compatible'' vehicles would merely be ``autonomous'' automated vehicles (AVs). In this case, such vehicles could yield occasionally (e.g. randomly) at a bottleneck and resume driving once the oncoming traffic flow ceases (in case of dense traffic: as soon as a vehicle on the blocked lane does not drive into the bottleneck, but instead decides to wait). This would contribute to more balanced traffic flow the more such AVs were around. It would, however, cause immediate disadvantages for any user of such systems, with no immediate advantage---and the disadvantage could be substantial in case other drivers on the blocked lane would take the opportunity and drive through the bottleneck in large numbers. Therefore, the waiting times for users of such an AV system could occasionally be very long and user acceptance would be highly questionable.

Do note that this model of the bottleneck is effectively symmetric in dense traffic: Vehicles on the blocked lane can only drive into the bottleneck if vehicles on the free lane yield the right of way---and vice versa. Traffic laws will typically state that the right of way primarily belongs to the vehicles on the free lane, but without strictly defined practical consequences. If a vehicle on the free lane yields, a (theoretically) arbitrary number of vehicles on the blocked lane may drain. We assume that in the non-connected variant, the free lane will typically flow at a much higher rate, because vehicles on the free lane will not usually yield, and vehicles on the blocked lane will not flow through the bottleneck in very large groups. This assumption is expressed in more rigid terms in the simulation evaluation in Sec.~\ref{sec:evaluation}.

\subsection{The ``I Want to Pass'' Variant}\label{sec:var-naive}

This variant would rely on vehicles on the blocked lane broadcasting some ``I want to pass'' message. Clearly, this signal provides little relevant information by itself: Any vehicle on the free lane may safely assume that any vehicle on the blocked lane (CAV or not) desires the right of way, so this request would be relatively redundant. Its only purpose could be that compatible CAVs on the free lane would identify the sender as a fellow system user, and therefore decide to yield to provide an advantage to the blocked compatible CAV.

This provides added value to the blocked CAV passengers; however, the CAV on the free lane would risk long waiting times just as in Sec.~\ref{sec:var-nonconnected}, in case other vehicles chose to drain as well. The usefulness of this variant over the variant described in Sec.~\ref{sec:var-nonconnected} is thus very limited.

\begin{framed}       
It can thus be concluded that the main challenge in bottleneck situations in dense, high-volume traffic is \emph{not} determining an appropriate opportunity to yield the right of way, but instead assuring acceptance by limiting the negative effects for any system user, as well as for the overall traffic flow.
\end{framed}

\subsection{The Proposed Variant}\label{sec:var-proposed}

\begin{figure}%
\includegraphics{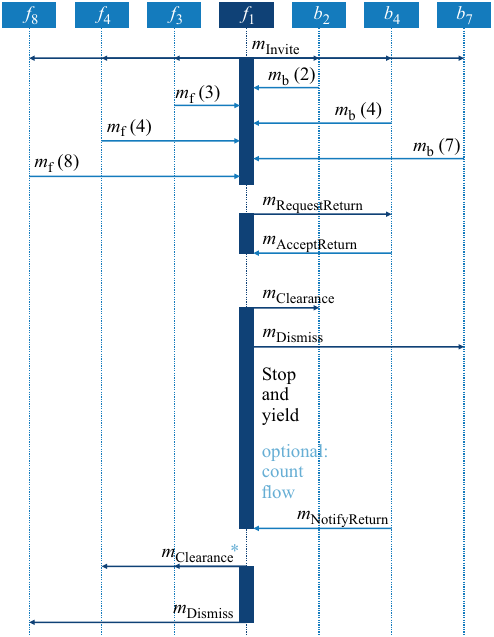}%
\caption{V2V messages exchanged at the example of Fig.~\ref{fig:example}, with non-connected vehicles omitted. The front CAV on the free lane, $f_1$, broadcasts $m\subs{Invite}$, and receives the positions $m\subs{b}$ and $m\subs{f}$ from the blocked and free CAVs respectively. $f_1$ identifies $b_4$ as the furthest admissible CAV based on the set $\tdmax = 5$, and requests $b_4$ to return the right of way via $m\subs{RequestReturn}$. Once $b_4$ confirms via $m\subs{AcceptReturn}$, $f_1$ allows $b_2$ to drain without stopping via $m\subs{Clearance}$, and informs $b_7$ that it will not drain in this round via $m\subs{Dismiss}$. $f_1$ then stops and yields, likely flashing the high beam light to communicate the yielding to human drivers. Optionally (cf. Sec.~\ref{sec:counting}) it counts draining vehicles (or measures elapsed time) until the opposite flow stops. Then $f_1$ drives into the bottleneck. In the non-counting variant, $f_3$ and $f_4$ will always receive permission to follow $f_1$ without yielding or sending $m\subs{Invite}$. In the counting variant (indicated by \textcolor{colorPRMedBlue}{$*$}), this will depend on the actions of the human drivers. If, for example, $b_3$ had decided to stop before the bottleneck, then only two vehicles from the blocked lane would have drained. In this case, $f_1$ would send $m\subs{Dismiss}$ to $f_3$ and $f_4$, advising them to offer $m\subs{Invite}$ before driving into the bottleneck. Either way, $f_8$ receives $m\subs{Dismiss}$.}%
\label{fig:diagrams}%
\end{figure}

This variant relies on the opposite principle to the one discussed in Sec.~\ref{sec:var-naive}: Vehicles on the blocked lane do not communicate their intention to pass the bottleneck, but instead their willingness to stop before entering it, and thereby cut off the flow from the blocked lane and return the right of way to the free lane. CAVs on the free lane can invite this cooperation---and only if such an ``inviter'' finds an appropriate ''returner'' on the blocked lane, the cooperation is actually executed.

It is assumed that any vehicle (both on the free lane and on the blocked lane) can be assigned a value $d$, denoting its distance from the bottleneck, its (estimated) time duration until passing the bottleneck, or the position in the queue before the bottleneck. The value of $d$ must be evaluated consistently and monotonously over queue positions by any CAV; hence, a metric distance (to be measured via GNSS, for example) can be practical. However, for the purpose of explanatory examples in this paper, when specific numbers are required we will assume $d = \td$, where $\td$ is the position in the queue, as shown in Fig.~\ref{fig:example}. The presented specific results are independent of the the particular choice of $d$ as long as it satisfies these requirements.

Each CAV on the free lane will have a parameter $\dmax$, likely picked by the human passengers, that denotes the maximum $d$ that the passengers of the inviter CAV are willing to accept to enable the cooperation: The inviter will allow any blocked vehicle $b_i$ to flow off whose $d_i \leq \dmax$. Based on this, we define the following V2V messages:

$m\subs{Invite}$: Sent by the front CAV on the free lane approaching the bottleneck (the ``inviter'') to invite other compatible CAVs to enter the cooperation.

$m\subs{b}(d), m\subs{f}(d)$: Sent by other CAVs (on the blocked, and on the free lane respectively) upon receiving $m\subs{Invite}$, to notify the inviter about their distance $d$.

$m\subs{RequestReturn}$: Sent by the inviter to the blocked CAV with the highest $d$ lower than the $\dmax+1$ of the inviter (the ``returner''). Thereby, the recipient is asked to stop before the bottleneck in case it reaches the front of the queue, and thereby return the right of way to the free lane after at most $\dmax$ vehicles have drained.

$m\subs{AcceptReturn}$: Sent by the returner to confirm the request. If this message is not received, the inviter will repeat the process with the CAV with the nearest lower $d$, if available.

$m\subs{NotifyReturn}$: Sent by the returner to notify that it has stopped in front of the bottleneck and returns the right of way to the free lane. This message is optional to increase efficiency. It can be omitted since the inviter requires the technical capability to recognize yielding (``returning'') human drivers anyway, however likely with additional delay.

$m\subs{Clearance}$: Sent by the inviter to another CAV to inform it that it may drive directly into the bottleneck without stopping or yielding. First, this message is sent to all blocked CAVs in front of the returner. Once the flow from the blocked lane ceases, $m\subs{Clearance}$ is sent to nearby rear CAVs on the free lane (up to a range of $\dmax$ of the inviter) to keep them from yielding and creating a bias in favor of the blocked lane.

$m\subs{Dismiss}$: Sent by the inviter to another CAV to inform it that any receiver (both on the free and on the blocked lane) is granted no permissions in this round. Any CAV on the blocked lane receiving this message must return to waiting for permission to drive into the bottleneck. Any receiving CAV on the free lane must send $m\subs{Invite}$ before entering the bottleneck. This message only serves as closure for other vehicles and can be omitted in practice, since no contradictory permissions are issued that would need to be revoked.

Therefore, the system design instead focuses on a CAD vehicle $b_i$ in the \emph{blocked} queue (the ``returner'') offering to stop before the bottleneck, if an oncoming \emph{free} CAD vehicle $f_1$ (the ``inviter'') lets all (human-driven or automated) vehicles $b_k,\; k < i$ (i.e. ahead of $b_i$) in the queue drain. This assures that the returner $b_i$ improves its position in the queue, while the inviter $f_1$ can rely on a limited maximum wait time, depending on the position of $b_i$ in the blocked queue. The returner $b_i$ is chosen based on the $\dmax$ of the inviter $f_1$ that can be picked by the passengers of the inviter. Thus, even for moderate penetrations of CAD vehicles, the returner is typically \emph{not} the blocked CAD vehicle closest to the bottleneck. This can allow several blocked CAVs (and other vehicles) to drain at each cycle, while any free CAV can limit its own waiting time. The negotiation is designed to require no centralized controller, but instead establishes balanced flow by emergence.

The developed human-machine interaction (HMI, elaborated in more detail in \cite{ehrhardt2021gap}) aims at communicating these particular effects to the passengers of the CAVs: Inviters on the free lane are provided a ``countdown'' metaphor that indicates the limited maximum waiting duration. The HMI for the returners on the blocked lane emphasizes the gains in queue progress due to cooperation.

\subsection{The counting vs. the non-counting variant}\label{sec:counting} We further distinguish the implementation into two specific variants. In the \emph{non-counting variant}, the inviter $f_1$ sends $m\subs{Clearance}$ to all subsequent vehicles on the free lane ($f_2, f_3, ...$) up to a distance of $\dmax$ of the inviter. Thereby the inviter will allow up to $\dmax$ free vehicles behind it to drain, regardless of how many blocked vehicles have drained in between. This variant is particularly easy to implement and provides a clearer benefit for users on the free lane; however, this variant will maintain a distinct bias in favor of the free lane (cf. Fig:~\ref{fig:diagrams} and Sec.~\ref{sec:counting-vs-noncounting}). The alternative implementation is the \emph{counting variant}, where the inviter $f_1$ counts the opposite flow while waiting to estimate the effective $d\subs{drained}$ of vehicles that have drained from the blocked lane. This may be implemented via actually counting passing vehicles, by measuring time, or by the ``returner'' communicating its travel distance until the blocked flow ceases. Then, the inviter sends $m\subs{Clearance}$ only to up to $d\subs{drained}$ subsequent CAVs on the free lane, to enforce a more balanced traffic flow. The two variants are compared in the evaluation in Sec.~\ref{sec:evaluation}.

\begin{figure}%
\includegraphics[width=\columnwidth]{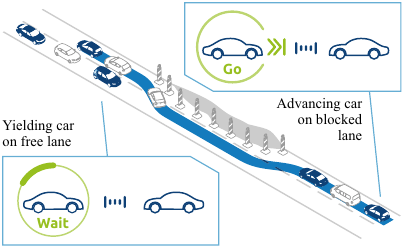}%
\caption{Example of the developed HMI based on \cite{ehrhardt2021gap}, with compatible CAVs shown as solid cars, and other vehicles shown as outlines. The front CAV on the left yields the right of way to oncoming vehicles on the blocked lane. To communicate the situation to the human passengers inside via an HMI, the connection to the partner vehicle is indicated via signal bars, and a countdown circle indicates the maximum waiting time. The partner vehicle, the CAV on the right, has agreed to stop before entering the bottleneck, and thus return the right of way. Its passengers are also shown a connection to the respective partner vehicle, but an arrow indicates the ``proceed until stop'' status. A study on the impact of such an HMI design is given in \cite{ehrhardt2021gap}.}%
\label{fig:hmi}%
\end{figure}

\section{Evaluation}\label{sec:evaluation}

\begin{figure}
\includegraphics[width=\columnwidth]{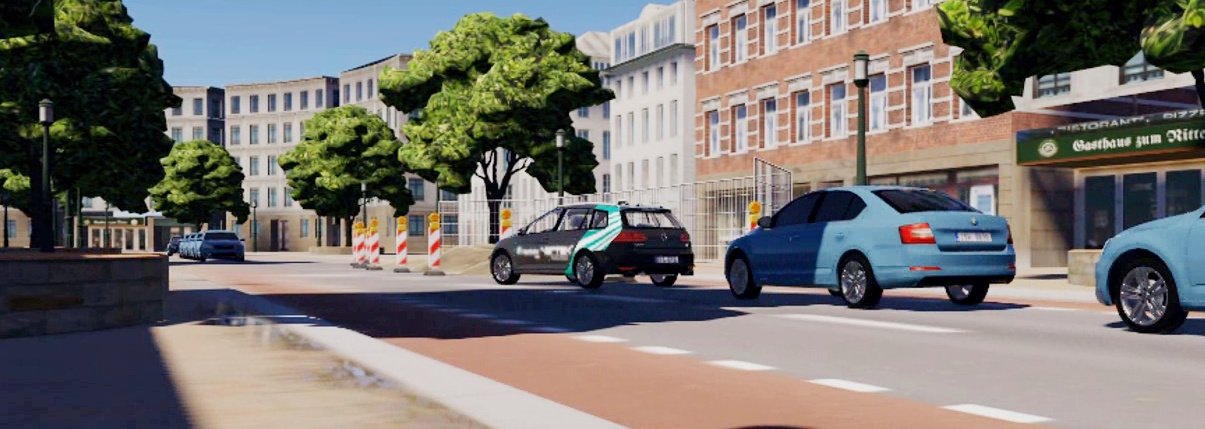}%
\caption[]{View of the scenario simulation in OCTANE\footnotemark{} showing a two-way street where one driving direction is blocked by a construction site. Vehicles in this direction have to drive on the opposite lane. In dense traffic, this requires an explicit stopping and yielding by the vehicles on the free lane.}%
\label{fig:octane}
\end{figure}
\footnotetext{\href{https://www.octane.org}{www.octane.org}}

\newcommand\carAny[1]{\colorbox{colorAny}{\!\textcolor{colorAny}{\texttt{00}}\!}}
\newcommand\carNum[1]{\colorbox{colorPRDGray}{\!\texttt{#1}\!}}
\newcommand\carFree[1]{\colorbox{colorPRMedBlue}{\!\texttt{\textcolor{white}{#1}}\!}}
\newcommand\carAgr[1]{\colorbox{colorPRDarkBlue}{\!\texttt{\textcolor{white}{#1}}\!}}

\newcommand\carAnyYield[1]{\colorbox{colorPRLightGreen}{\!\textcolor{colorPRLightGreen}{\texttt{00}}\!}}
\newcommand\carNumYield[1]{\colorbox{colorPRLightGreen}{\!\texttt{#1}\!}}

\newcommand\driveLeft{\quad\quad $\blacktriangleleft$\quad }
\newcommand\driveRight{\quad $\blacktriangleright$\quad\quad }
\newcommand\newlineCarPlot{\\[2pt]}

\newcounter{carplotline}

\newcommand\carPlotLine{\refstepcounter{carplotline}\noindent \llap{$\tau = \thecarplotline$}\; }

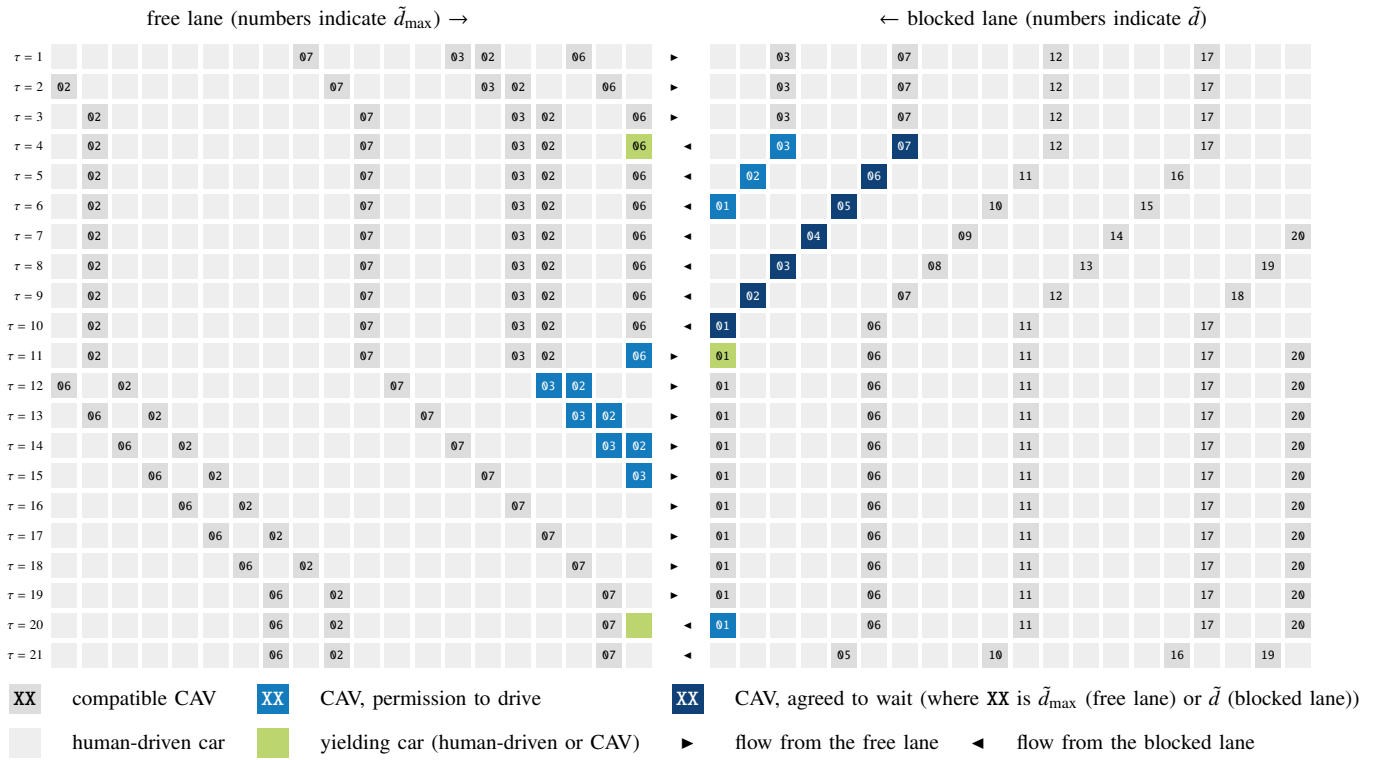
\begin{figure*}%
\footnotesize
\begin{center}
free lane (numbers indicate $\tdmax$) $\rightarrow$\hspace{0.3\textwidth}$\leftarrow$ blocked lane (numbers indicate $\tilde{d}$)
\end{center}
\begin{center}
\tiny
\input{carplot}
\end{center}
\begin{tabular}{clclclcl}
\footnotesize
\carNum{XX}& compatible CAV &
\carFree{XX}& CAV, permission to drive&
\carAgr{XX}& \multicolumn{3}{l}{CAV, agreed to wait (where \texttt{XX} is $\tdmax$ (free lane) or $\tilde{d}$ (blocked lane))}\\[5pt]
\carAny{00}& human-driven car&
\carAnyYield{00}& yielding car (human-driven or CAV)&
$\blacktriangleright$& flow from the free lane&
$\blacktriangleleft$& flow from the blocked lane
\end{tabular}
\caption{Extract of the simulation scenarios from a turn-based perspective of vehicle movements sliced by relevant actions ($\tau$) instead of time. Vehicles which have already passed the bottleneck are not shown. Numbered vehicles are compatible CAVs. As a measure of distance, $\tilde{d}$, the approximate number of cars to the bottleneck, is used and indicated. For vehicles on the free lane (left) their choice for $\tdmax$ is given. For vehicles on the blocked lane (right) the $\tilde{d}$ corresponding to their current position is indicated. Any unnumbered box is a vehicle that acts based on stochastic human behavior parameters.\\[0.5em]
The scenario shows a compatible CAV yielding the right of way at $\tau = 3$ based on the acceptance of the blocked CAV at $\tilde{d} = 7$ to wait. This benefits the CAV in front of it ($\tilde{d} = 3$) which receives explicit permission to drive into the bottleneck. The rear vehicle waits as promised at $\tau = 10$ and returns the right of way to the free lane. It is granted the right of way by an obliging human driver at $\tau = 20$ before the next CAV on the free lane arrives at the front (which would also have yielded, based on its $\tdmax = 7$ and the availability of a blocked CAV at $\tilde{d}=6$.}%
\label{fig:scenario}%
\end{figure*}

\begin{figure*}%
\includegraphics{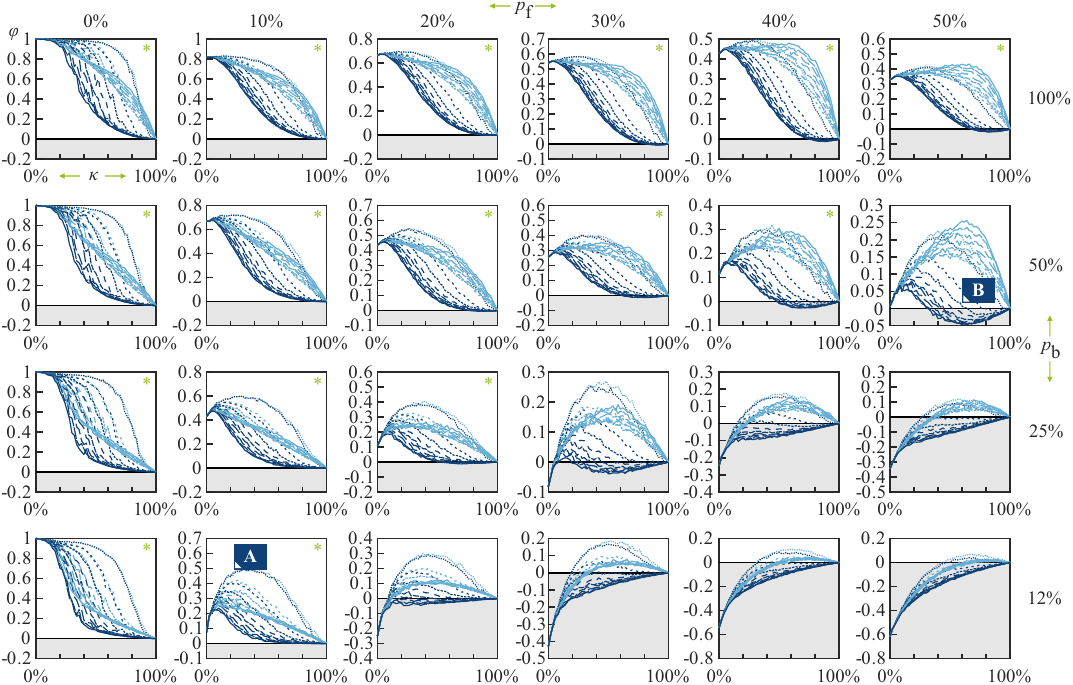}%
\caption{Traffic flow $\varphi$ over different percentages of compatible CAVs $\kappa$, by various parameters of human behavior (the probability of a human driver to wait before driving into the bottleneck from the free lane $p_{\text{f}}$ and from the blocked lane $p_{\text{b}}$). More solid lines indicate higher $\tdmaxmax$, light lines denote the non-counting variant, dark lines denote the counting variant. The shaded area denotes reversed flow bias, with more vehicles on the blocked lane draining at this parameter combination. More ``likely'' parameter combinations, namely where human drivers on the free lane are more assertive than drivers on the blocked lane ($p\subs{f} < p\subs{b}$), are marked with an \textcolor{colorPRGreen}{$*$}. These parameter combinations are revisited in accumulated form in Fig.~\ref{fig:simparameterresults-accumulated}.}%
\label{fig:simparameterresults}%
\end{figure*}

The system was evaluated in a traffic simulation with different parameters for human behavior, different CAV algorithm designs, and different penetrations of compatible CAVs among other traffic participants. In each case, the scenario was a two-way street with one lane in each direction (cf. Fig.~\ref{fig:octane})---one blocked by a construction site, the other free. Traffic volume was taken to be so high that negotiation between vehicles on opposite sides of the bottleneck was always required, to assure that the negotiation algorithm is always active in the results. The simulation only differentiates between compatible CAVs and ``other'' traffic participants, with no distinction between human drivers and non-cooperative AVs.

To reduce the number of free parameters and required assumptions, the simulation is viewed from a turn-based perspective: Every vehicle to drain through the bottleneck is taken as a ``turn'', as well as any change flow direction. This leads to the simplified view of the simulation that is shown in Fig.~\ref{fig:scenario}, where effects of different models of car-following behavior (such as Wiedemann \cite{wiedemann1974simulation, fellendorf2010microscopic} or Krauss \cite{krauss1998microscopic, krajzewicz2002sumo}) do not have to be considered, and neither does6 the local speed limit or the length of the bottleneck. As a result (and limitation), actual driving and waiting times, which depend significantly on these parameters, cannot be compared\footnote{For example, frequent changes in driving directions, at slow driving speeds, with slowly-reacting drivers driving through a long bottleneck, will lead to considerably longer turn times since any change in direction requires all vehicles to clear the bottleneck first.}; instead, we consider a fixed number of 50\,000 ``turns'' per combination of parameters and evaluate the balance in traffic flow, which under the given practical conditions corresponds to a simulated duration around three days of constant traffic per scenario (depending on the parameters). Yet, this particular simulation setup allows to evaluate an envelope over a wide range of plausible scenarios designed such that a majority of real, complex scenarios should approximately and stochastically be covered, providing a firmer empirical basis than small-scale real-world experiments could.

\subsection{Simulation Parameters}

\subsubsection{CAV setup} The simulated behavior of the CAVs is based on the proposed algorithmic implementation, with the non-counting and the counting variant compared in different runs. Based on an analysis of plausible V2V ranges in urban scenarios (details cf. \cite{Kowalewski2020_1000099791}), a negotiation distance of up to 20 vehicles in each direction was considered, which corresponds to a distance of around 150\,m or 400 feet plus the length of the bottleneck, and the communication over this distance is assumed to be sufficiently stable to robustly exchange the limited amount of message data in the proposed implementation. CAVs are assumed to have their individual parameter $\tdmax$ set by a human passenger based on a discrete uniform distribution with parameter $\tdmaxmax$, such that
\begin{equation}
\begin{aligned}
\tdmax &\sim \mathcal{U}_{\mathbb N}(1, \tdmaxmax)\quad\text{and}\\
\tdmaxmax &\in \{4, 6, 8, ..., 20\}.
\end{aligned}
\label{eq:}
\end{equation}
By these assumptions, the high traffic volume and the turn-based perspective, this algorithm is always active and determines the behavior of the CAV completely (w.r.t. the results).

\subsubsection{Non-CAV behavior} Any vehicle that is not a compatible CAV follows a stochastic behavior model. This is taken to be a fair representation of human drivers (since human behavior varies strongly and datasets of human behavior are scarce) as well as many potential implementations for any non-compatible kind of connected or non-connected automated driving. The behavior model distinguishes between the probability $p\subs{f}$ of a vehicle on the free lane yielding the right of way to the blocked lane, and the probability $p\subs{b}$ of a blocked vehicle returning the right of way (instead of following its predecessor into the bottleneck). It is assumed that, at each such action, the \emph{first} vehicle on the opposite lane \emph{always} drives into the bottleneck (since it received explicit permission, e.g. by flashing headlights). Subsequent non-CAVs decide based on $p\subs{f}$ or $p\subs{b}$ respectively. The evaluated parameters (as shown in Fig.~\ref{fig:simparameterresults}) are
\begin{equation}
\begin{aligned}
p\subs{f} &\in \{0\,\%, 10\,\%, 20\,\%, 30\,\%, 40\,\%, 50\,\%\}\quad\text{and}\\
p\subs{b} &\in \{12\,\%, 25\,\%, 50\,\%, 100\,\%\}
\end{aligned}
\label{eq:}
\end{equation}

\subsubsection{Overall simulation} To determine the effect of different penetrations of compatible CAVs compared to non-compatible vehicles, different ratios $\kappa = (\text{number of compatible CAVs})/(\text{number of all vehicles})$ were evaluated, with
\begin{equation}
\kappa \in \{0\,\%, 2\,\%, 4\,\%, ..., 98\,\%, 100\,\%\},
\end{equation}
where $\kappa = 0\,\%$ represents current-day traffic without compatible CAVs. This leads to an overall parameter sample count of $N_{\tdmaxmax} \times N_{p\subs{f}} \times N_{p\subs{b}} \times N_\kappa = 11\,016$, each of which was evaluated over a duration of 50\,000 turns.

Each scenario is evaluated via the flow balance metric
\begin{equation}
\varphi = 2 \cdot \frac{\text{number of drained vehicles on free lane}}{\text{number of all vehicles}} - 1
\label{eq:}
\end{equation}
such that $\varphi = 1$ denotes that only vehicles from the free lane drain, $\varphi = 0$ denotes a perfectly balanced flow, and $\varphi = -1$ denotes the case that only vehicles on the blocked lane drain.

\subsection{Results}

The results of this evaluation are shown in Fig.~\ref{fig:simparameterresults} (distinguished by all parameters) and Fig.~\ref{fig:simparameterresults-accumulated} (accumulated over ``likely'' parameters). They show various notable effects that will be discussed in the following.

Firstly, each variant achieves the desired result that $\varphi \rightarrow 0$ as $\kappa \rightarrow 100\,\%$, so the algorithm does support an evolutionary introduction of compatible CAVs. The effect on overall traffic flow depends significantly on the parameters of the human drivers. For low $p\subs{f}$ and relatively high $p\subs{b}$ (expected to be more common in current traffic), the impact of the proposed CAD function is significant especially for vehicles on the blocked lane (which gain approximately $1\,\%$ improved $\varphi$ for each $1\,\%$ increase in $\kappa > 10\,\%$). For $p\subs{f} \approx p\subs{b}$, however, due to the symmetry in the original model (which is the basis for the stochastic human behavior), a balance of $\varphi\approx 0$ is already achieved at $\kappa=0$.

For $p\subs{f} \gg p\subs{b}$, i.e. very obliging human drivers on the free lane, and very assertive human drivers on the blocked lane, traffic flow is reversed to $\varphi < 0$ at $\kappa = 0$, and blocked vehicles drain in greater numbers than free vehicles. Introduction of compatible CAVs again resolves this situation, though not as effectively as with the more likely case of assertive drivers on the free lane and careful drivers on the blocked lane.

Several notable effects stand out where the introduction of compatible CAVs has counterintuitive consequences.

\subsubsection{Late onset of effects for $p\subs{f} = 0\,\%$ and low $p\subs{b}$} In this case, traffic flow on the blocked lane effectively ceases. Human drivers on the free lane never yield, CAVs on the free lane rarely ever yield, and even if they do, blocked vehicles hardly drain; therefore there is little to no progress on the blocked lane, unless sufficiently many connected CAVs on both lanes are available.

\subsubsection{Imbalance introduced by CAVs}\label{sec:effect-imbalance} With increasingly assertive blocked vehicles (low $p\subs{b}$), the introduction of CAVs increases an original imbalance for $\varphi > 0$. For example, an almost balanced flow ($p\subs{f} = 10\,\%, p\subs{b} = 12\,\%$; marker \textbf{A} in Fig.~\ref{fig:simparameterresults}) of $\varphi = 0.08$ increases to $\varphi > 0.22$ at $\kappa = 10\,\%$. This exposes the asymmetry introduced by the proposed algorithm. The algorithm primarily imposes a sharp limit of waiting time for compatible CAVs on the free lane, and only secondarily balances traffic flow. Hence, any condition where a balanced flow would require very long waiting times contradicts these priorities. In the above example, typically groups of 8 cars or more would pass the bottleneck at $\kappa = 0$. Maintaining this would require a sufficiently high $\tdmax$ for each CAV. If CAVs with $\tdmax = 2$ (for example) dominate, then these CAVs would not contribute to achieving this balance. In this case, CAV passengers on the free lane benefit from reduced waiting times but overall traffic flow would be imbalanced by impatient CAV passengers.

\subsubsection{Reversed flow bias by CAVs}\label{sec:effect-reversed-flow} High numbers of compatible CAVs can bias the flow at $\varphi > 0$ (for $\kappa=0$) towards $\varphi < 0$ (for $\kappa > 0$), as seen at marker \textbf{B} in Fig.~\ref{fig:simparameterresults}, for more obliging human drivers on the free lane. Here, the permission granted to subsequent CAVs on the free lane can be interrupted by human drivers in between, who return the right of way to the blocked lane. In this case, the free lane is at a slight advantage when enough CAVs are present to enable the algorithm, but obliging human drivers on the free lane interrupt the drain of CAVs which received the explicit permission.

Due to the stochastic nature of the simulation, both effects in Secs. \ref{sec:effect-imbalance} and \ref{sec:effect-reversed-flow} occur at any parameter combination, but for expected realistic conditions $p\subs{f} \ll p\subs{b}$, their impact is slight compared to the uncertainty of human behavior.

\subsubsection{Counting vs. non-counting variant}\label{sec:counting-vs-noncounting} 
\begin{figure}%
\centering
\includegraphics{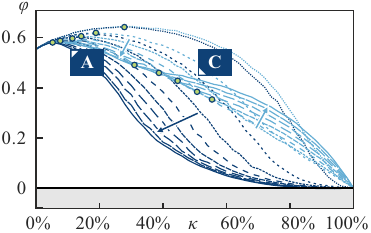}%
\caption{Accumulation of simulation results over more ``likely'' scenarios, namely those in Fig.~\ref{fig:simparameterresults} marked with an \textcolor{colorPRGreen}{$*$} where $p\subs{f} < p\subs{b}$. Arrows indicate trends by increasing $\tdmaxmax$. Typical effects are seen more clearly: Firstly, the transition at \textbf{A} from an increase in $\varphi$ to a decrease, which occurs at lower $\kappa$ in cases of higher $\tdmaxmax$. Secondly, the transition at \textbf{C} from a strong balancing effect of increased $\tdmaxmax$ towards a much weaker effect, once enough compatible CAVs are available that $\tdmaxmax$ does not notably increase the likelihood of a match anymore, but instead primarily increases the number of cars draining from the free lane. In contrast to the less likely scenarios in Fig.~\ref{fig:simparameterresults} (where  $p\subs{f} > p\subs{b}$), these likely scenarios do not reverse the flow to put vehicles on the blocked lane at an advantage.}%
\label{fig:simparameterresults-accumulated}%
\end{figure}

In comparison between the counting vs. the non-counting variant, we find that, as expected, the non-counting variant is much less effective at reducing the flow bias towards $\varphi\approx0$, since yielding CAVs on the free lane usually invite more subsequent vehicles to follow them than blocked vehicles have drained in between. The effect, denoted \textbf{C} in Fig.~\ref{fig:simparameterresults-accumulated}, is primarily increased by more careful drivers in the blocked lane: For $p\subs{b} = 50\,\%$, for instance, the number of blocked cars draining per turn is typically around 2 for low to moderate $\kappa$, regardless of $\tdmax$ of the yielding CAV. The number of cars subsequently draining from the free lane is then increased by the choice of $\tdmax$, and reduced only if human drivers in between decide to yield earlier (which more likely at higher $p\subs{f}$). This leads to the effect that at high penetrations of compatible CAVs and high $p\subs{b}$, a higher $\tdmax$ loses the effect of letting more oncoming vehicles drain, and primarily serves to benefit more subsequent vehicles on the free lane---even if only one blocked vehicle has drained. In this case (as best seen in Fig.~\ref{fig:simparameterresults-accumulated}), the reversal occurs already at lower $\kappa$ for higher $\tdmax$, because the likelihood of a match increases.

\section{Conclusion and Outlook}\label{sec:conclusion}

We have presented a basic algorithmic principle to resolve bottleneck scenarios by cooperative decision making in connected automated vehicles (CAVs) in high-volume, dense traffic, that can act as a lean additional system to address this rather specific kind of scenario. The proposed solution achieves low requirements on communication and computation technology by leveraging emergence instead of global planning, and provides an intuitive principle that can be communicated to users with relative ease. The design takes technological, traffic-related and human-factor related aspects into account, and specifically aims at providing a solution that supports all levels of penetration ($\kappa$) of compatible CAVs in traffic: From the first few such vehicles, over an incremental (evolutionary) increase, up to a full penetration---and beyond, by providing an upward-compatible function that facilitates the introduction of later generations of CAVs (and thus in turn the future, evolutionary replacement of the proposed systems).

Two proposed variants have been evaluated separately: a \emph{non-counting} variant, where vehicles on the free lane act independently of the amount of blocked vehicles actually draining (vs. hesitating); and a \emph{counting} variant that reduces flow on the free lane to match the actual flow on the blocked lane. For each variant, a total of $11\,016$ parameter combinations as evaluated over $50\,000$ ``turns'' in traffic flow each.

The results show that, over a wide range of possible future scenarios (including stochastic behavior parameters of human drivers or non-cooperative AVs, penetrations of compatible CAVs, and algorithm parameter choices), the proposed solution robustly approaches an equilibrium in traffic flow for increasing penetrations of compatible CAVs, $\kappa$.  As expected, the counting variant is more effective in reaching a balanced traffic flow; however, uncertainties in technical and behavioral details make it impossible to definitively compare gains (i.e. improved traffic balance, improved perceived ``fairness'') and costs (i.e. more complex system design, potentially lower overall traffic efficiency due to more frequent direction changes). However, the system's quantitative performance is not uniform over all considered parameter values (and parameter combinations), and can in some cases exhibit counterintuitive artifacts. Over all considered parameter values, none of these artifacts had dramatic effects on the overall traffic flow, but depending on an evaluation of likelihood of certain parameter combinations (which is beyond the scope of this paper), dedicated extensions to the basic principle may be desirable that compensate these effects.

The results indicate that an evolutionary introduction of advanced CAD functions is possible. In this way, systems as presented here may serve to bridge the gap between close-to-market technologies and low peneration rates, and future technologies / high penetration rates, by providing added value both to users and overall traffic flow, and by being compatible both to stochastic (human) behavior, and by providing a clear and generic interface that can be considered and leveraged by future CAVs that will supersede the proposed systems.

Since the performance of such systems is significantly affected by technological feasibility and efficiency, traffic impact, human factors, as well as acceptance and market value, solutions for complex CAV systems must express goals and requirements from all of these perspectives to develop a robust solution that is widely adopted. For this reason, it is anticipated that a broad, interdisciplinary approach will be pivotal in system design and evaluation.

\subsection*{Outlook}

The presented model has introduced several simplifying assumptions to make the presentation and evaluation practical, and to focus on common basic principles and effects. Thereby, the results and the conclusions drawn therefrom are claimed to be accurate, but expose performance only under a specific set of scenarios and with reference to abstract performance metrics (e.g. balance of vehicle flow numbers and ``turns'', as opposed to waiting times, emissions, etc.).

For example, the proposed system is only meaningfully defined in high-volume, dense traffic. A CAD solution for low-volume, sparse traffic is presented in \cite{naumann2017cooperativePlanning}, and an intermediate study on human factors is given in \cite{ehrhardt2021gap}. A generic solution must be able to also scale well with different flow volumes, interpolating between these individual scenarios.

On the basis of more detailed technical design parameters, and more available data on human traffic behavior, a transition from the turn-based evaluation perspective towards a substantiated cost estimate (in terms of waiting times, technological resources) would be possible and in fact necessary to distinguish between different design variants and parametrizations that appear similar in performance in the vehicle-based, turn-based perspective.

The presented model further assumes that a balanced flow in both directions represents the optimal solution. If, however, traffic demand is notably different in each direction (such as typically during rush hours), this should be included in the model, such that the high-demand direction is favored and the equilibrium is shifted to a ratio that corresponds to the ratio in traffic demands. Thereby the waiting time accumulated over traffic participants would be balanced instead of the local flow at the bottleneck. Thus, a relevant future extension lies in including variable traffic demand by direction into the model, as well as different penetrations of CAVs per direction, and defining a suitable tradeoff model between accumulated and individual waiting time.

In the paper, the bottleneck is modeled as stationary over the considered time intervals. For some obstacles, such as garbage collection or slow tractors, bottlenecks move, albeit slowly. In high-volume traffic, this means among other things that yielding requires more foresight, and traffic flow on the blocked lane is not actually zero while the free lane is flowing. Hence, one relevant extension of the model lies in including dynamic bottlenecks and correspondingly adapted optimality criteria.

Furthermore, and very importantly, the paper intentionally does not specify any incentive model to promote the continuous use of such a system---in particular by drivers on the free lane. The presented approach focuses on establishing a basic principle that ensures that disadvantages to any users are strictly limited (with respect to the non-cooperative scenario). Preliminary results from ongoing research in the context of the project suggest that a majority of potential users would be willing to sacrifice driving time purely for the benefit of overall traffic flow \emph{if} their own maximum disadvantage could be limited in such a way; however, it is unclear how reliable these predictions are in future everyday scenarios. It appears likely that, in practice, concrete benefits could significantly improve acceptance and widespread use, and thereby effectiveness. Since efficient traffic flow is a public benefit, incentive models may for example include funded discounts on refueling / charging for cooperative users. The definition of a possible maximum ``disadvantage'' or ``cost'' mechanism, provided in this paper, is intended as a concrete basis for gauging the options and requirements for a corresponding incentive system.

\IEEEpeerreviewmaketitle

\bibliographystyle{alpha}
\bibliography{bottleneck}

\end{document}

%% file: includes.tex
\usepackage{amsmath}
\usepackage{amssymb}
\usepackage{units}
\usepackage[hidelinks]{hyperref}

\usepackage[table]{xcolor}

\usepackage{braket}
\usepackage{tikz}
\usetikzlibrary{arrows}
\usepackage{algorithm2e}
\usepackage{array}
\usepackage{framed, color}
\usepackage[normalem]{ulem}
\usepackage{cancel}
\usepackage{overpic}
\usepackage[footnotesize]{caption}
\usepackage[font+=footnotesize]{subcaption}
\usepackage[nodisplayskipstretch]{setspace}
\usepackage{txfonts}
\let\mathbb=\varmathbb

\usepackage{url}
\usepackage{setspace}
\usepackage{units}
\usepackage{upgreek}
\usepackage[textwidth=1.5cm]{todonotes}
\usepackage{dblfloatfix}

\usepackage[a4paper,left=1.5cm,right=1.5cm,top=2cm,bottom=2cm]{geometry}

\newcommand{\subs}[1]{_{\textrm{\upshape{#1}}}}

\definecolor{colorFHG}{rgb}{0.09,0.61,0.49}
\definecolor{colorPenalty}{rgb}{0.94,0.27,0.05}
\definecolor{colorPRGreen}{HTML}{97BF20}

%% file: carplot.tex
 \carPlotLine{}\carAny{--} \carAny{--} \carAny{--} \carAny{--} \carAny{--} \carAny{--} \carAny{--} \carAny{--} \carNum{07} \carAny{--} \carAny{--} \carAny{--} \carAny{--} \carNum{03} \carNum{02} \carAny{--} \carAny{--} \carNum{06} \carAny{--} \carAny{--} \driveRight{} \carAny{--} \carAny{--} \carNum{03} \carAny{--} \carAny{--} \carAny{--} \carNum{07} \carAny{--} \carAny{--} \carAny{--} \carAny{--} \carNum{12} \carAny{--} \carAny{--} \carAny{--} \carAny{--} \carNum{17} \carAny{--} \carAny{--} \carAny{--} \newlineCarPlot
 \carPlotLine{}\carNum{02} \carAny{--} \carAny{--} \carAny{--} \carAny{--} \carAny{--} \carAny{--} \carAny{--} \carAny{--} \carNum{07} \carAny{--} \carAny{--} \carAny{--} \carAny{--} \carNum{03} \carNum{02} \carAny{--} \carAny{--} \carNum{06} \carAny{--} \driveRight{} \carAny{--} \carAny{--} \carNum{03} \carAny{--} \carAny{--} \carAny{--} \carNum{07} \carAny{--} \carAny{--} \carAny{--} \carAny{--} \carNum{12} \carAny{--} \carAny{--} \carAny{--} \carAny{--} \carNum{17} \carAny{--} \carAny{--} \carAny{--} \newlineCarPlot
 \carPlotLine{}\carAny{--} \carNum{02} \carAny{--} \carAny{--} \carAny{--} \carAny{--} \carAny{--} \carAny{--} \carAny{--} \carAny{--} \carNum{07} \carAny{--} \carAny{--} \carAny{--} \carAny{--} \carNum{03} \carNum{02} \carAny{--} \carAny{--} \carNum{06} \driveRight{} \carAny{--} \carAny{--} \carNum{03} \carAny{--} \carAny{--} \carAny{--} \carNum{07} \carAny{--} \carAny{--} \carAny{--} \carAny{--} \carNum{12} \carAny{--} \carAny{--} \carAny{--} \carAny{--} \carNum{17} \carAny{--} \carAny{--} \carAny{--} \newlineCarPlot
 \carPlotLine{}\carAny{--} \carNum{02} \carAny{--} \carAny{--} \carAny{--} \carAny{--} \carAny{--} \carAny{--} \carAny{--} \carAny{--} \carNum{07} \carAny{--} \carAny{--} \carAny{--} \carAny{--} \carNum{03} \carNum{02} \carAny{--} \carAny{--} \carNumYield{06} \driveLeft{} \carAny{--} \carAny{--} \carFree{03} \carAny{--} \carAny{--} \carAny{--} \carAgr{07} \carAny{--} \carAny{--} \carAny{--} \carAny{--} \carNum{12} \carAny{--} \carAny{--} \carAny{--} \carAny{--} \carNum{17} \carAny{--} \carAny{--} \carAny{--} \newlineCarPlot
 \carPlotLine{}\carAny{--} \carNum{02} \carAny{--} \carAny{--} \carAny{--} \carAny{--} \carAny{--} \carAny{--} \carAny{--} \carAny{--} \carNum{07} \carAny{--} \carAny{--} \carAny{--} \carAny{--} \carNum{03} \carNum{02} \carAny{--} \carAny{--} \carNum{06} \driveLeft{} \carAny{--} \carFree{02} \carAny{--} \carAny{--} \carAny{--} \carAgr{06} \carAny{--} \carAny{--} \carAny{--} \carAny{--} \carNum{11} \carAny{--} \carAny{--} \carAny{--} \carAny{--} \carNum{16} \carAny{--} \carAny{--} \carAny{--} \carAny{--} \newlineCarPlot
 \carPlotLine{}\carAny{--} \carNum{02} \carAny{--} \carAny{--} \carAny{--} \carAny{--} \carAny{--} \carAny{--} \carAny{--} \carAny{--} \carNum{07} \carAny{--} \carAny{--} \carAny{--} \carAny{--} \carNum{03} \carNum{02} \carAny{--} \carAny{--} \carNum{06} \driveLeft{} \carFree{01} \carAny{--} \carAny{--} \carAny{--} \carAgr{05} \carAny{--} \carAny{--} \carAny{--} \carAny{--} \carNum{10} \carAny{--} \carAny{--} \carAny{--} \carAny{--} \carNum{15} \carAny{--} \carAny{--} \carAny{--} \carAny{--} \carAny{--} \newlineCarPlot
 \carPlotLine{}\carAny{--} \carNum{02} \carAny{--} \carAny{--} \carAny{--} \carAny{--} \carAny{--} \carAny{--} \carAny{--} \carAny{--} \carNum{07} \carAny{--} \carAny{--} \carAny{--} \carAny{--} \carNum{03} \carNum{02} \carAny{--} \carAny{--} \carNum{06} \driveLeft{} \carAny{--} \carAny{--} \carAny{--} \carAgr{04} \carAny{--} \carAny{--} \carAny{--} \carAny{--} \carNum{09} \carAny{--} \carAny{--} \carAny{--} \carAny{--} \carNum{14} \carAny{--} \carAny{--} \carAny{--} \carAny{--} \carAny{--} \carNum{20} \newlineCarPlot
 \carPlotLine{}\carAny{--} \carNum{02} \carAny{--} \carAny{--} \carAny{--} \carAny{--} \carAny{--} \carAny{--} \carAny{--} \carAny{--} \carNum{07} \carAny{--} \carAny{--} \carAny{--} \carAny{--} \carNum{03} \carNum{02} \carAny{--} \carAny{--} \carNum{06} \driveLeft{} \carAny{--} \carAny{--} \carAgr{03} \carAny{--} \carAny{--} \carAny{--} \carAny{--} \carNum{08} \carAny{--} \carAny{--} \carAny{--} \carAny{--} \carNum{13} \carAny{--} \carAny{--} \carAny{--} \carAny{--} \carAny{--} \carNum{19} \carAny{--} \newlineCarPlot
 \carPlotLine{}\carAny{--} \carNum{02} \carAny{--} \carAny{--} \carAny{--} \carAny{--} \carAny{--} \carAny{--} \carAny{--} \carAny{--} \carNum{07} \carAny{--} \carAny{--} \carAny{--} \carAny{--} \carNum{03} \carNum{02} \carAny{--} \carAny{--} \carNum{06} \driveLeft{} \carAny{--} \carAgr{02} \carAny{--} \carAny{--} \carAny{--} \carAny{--} \carNum{07} \carAny{--} \carAny{--} \carAny{--} \carAny{--} \carNum{12} \carAny{--} \carAny{--} \carAny{--} \carAny{--} \carAny{--} \carNum{18} \carAny{--} \carAny{--} \newlineCarPlot
 \carPlotLine{}\carAny{--} \carNum{02} \carAny{--} \carAny{--} \carAny{--} \carAny{--} \carAny{--} \carAny{--} \carAny{--} \carAny{--} \carNum{07} \carAny{--} \carAny{--} \carAny{--} \carAny{--} \carNum{03} \carNum{02} \carAny{--} \carAny{--} \carNum{06} \driveLeft{} \carAgr{01} \carAny{--} \carAny{--} \carAny{--} \carAny{--} \carNum{06} \carAny{--} \carAny{--} \carAny{--} \carAny{--} \carNum{11} \carAny{--} \carAny{--} \carAny{--} \carAny{--} \carAny{--} \carNum{17} \carAny{--} \carAny{--} \carAny{--} \newlineCarPlot
 \carPlotLine{}\carAny{--} \carNum{02} \carAny{--} \carAny{--} \carAny{--} \carAny{--} \carAny{--} \carAny{--} \carAny{--} \carAny{--} \carNum{07} \carAny{--} \carAny{--} \carAny{--} \carAny{--} \carNum{03} \carNum{02} \carAny{--} \carAny{--} \carFree{06} \driveRight{} \carNumYield{01} \carAny{--} \carAny{--} \carAny{--} \carAny{--} \carNum{06} \carAny{--} \carAny{--} \carAny{--} \carAny{--} \carNum{11} \carAny{--} \carAny{--} \carAny{--} \carAny{--} \carAny{--} \carNum{17} \carAny{--} \carAny{--} \carNum{20} \newlineCarPlot
 \carPlotLine{}\carNum{06} \carAny{--} \carNum{02} \carAny{--} \carAny{--} \carAny{--} \carAny{--} \carAny{--} \carAny{--} \carAny{--} \carAny{--} \carNum{07} \carAny{--} \carAny{--} \carAny{--} \carAny{--} \carFree{03} \carFree{02} \carAny{--} \carAny{--} \driveRight{} \carNum{01} \carAny{--} \carAny{--} \carAny{--} \carAny{--} \carNum{06} \carAny{--} \carAny{--} \carAny{--} \carAny{--} \carNum{11} \carAny{--} \carAny{--} \carAny{--} \carAny{--} \carAny{--} \carNum{17} \carAny{--} \carAny{--} \carNum{20} \newlineCarPlot
 \carPlotLine{}\carAny{--} \carNum{06} \carAny{--} \carNum{02} \carAny{--} \carAny{--} \carAny{--} \carAny{--} \carAny{--} \carAny{--} \carAny{--} \carAny{--} \carNum{07} \carAny{--} \carAny{--} \carAny{--} \carAny{--} \carFree{03} \carFree{02} \carAny{--} \driveRight{} \carNum{01} \carAny{--} \carAny{--} \carAny{--} \carAny{--} \carNum{06} \carAny{--} \carAny{--} \carAny{--} \carAny{--} \carNum{11} \carAny{--} \carAny{--} \carAny{--} \carAny{--} \carAny{--} \carNum{17} \carAny{--} \carAny{--} \carNum{20} \newlineCarPlot
 \carPlotLine{}\carAny{--} \carAny{--} \carNum{06} \carAny{--} \carNum{02} \carAny{--} \carAny{--} \carAny{--} \carAny{--} \carAny{--} \carAny{--} \carAny{--} \carAny{--} \carNum{07} \carAny{--} \carAny{--} \carAny{--} \carAny{--} \carFree{03} \carFree{02} \driveRight{} \carNum{01} \carAny{--} \carAny{--} \carAny{--} \carAny{--} \carNum{06} \carAny{--} \carAny{--} \carAny{--} \carAny{--} \carNum{11} \carAny{--} \carAny{--} \carAny{--} \carAny{--} \carAny{--} \carNum{17} \carAny{--} \carAny{--} \carNum{20} \newlineCarPlot
 \carPlotLine{}\carAny{--} \carAny{--} \carAny{--} \carNum{06} \carAny{--} \carNum{02} \carAny{--} \carAny{--} \carAny{--} \carAny{--} \carAny{--} \carAny{--} \carAny{--} \carAny{--} \carNum{07} \carAny{--} \carAny{--} \carAny{--} \carAny{--} \carFree{03} \driveRight{} \carNum{01} \carAny{--} \carAny{--} \carAny{--} \carAny{--} \carNum{06} \carAny{--} \carAny{--} \carAny{--} \carAny{--} \carNum{11} \carAny{--} \carAny{--} \carAny{--} \carAny{--} \carAny{--} \carNum{17} \carAny{--} \carAny{--} \carNum{20} \newlineCarPlot
 \carPlotLine{}\carAny{--} \carAny{--} \carAny{--} \carAny{--} \carNum{06} \carAny{--} \carNum{02} \carAny{--} \carAny{--} \carAny{--} \carAny{--} \carAny{--} \carAny{--} \carAny{--} \carAny{--} \carNum{07} \carAny{--} \carAny{--} \carAny{--} \carAny{--} \driveRight{} \carNum{01} \carAny{--} \carAny{--} \carAny{--} \carAny{--} \carNum{06} \carAny{--} \carAny{--} \carAny{--} \carAny{--} \carNum{11} \carAny{--} \carAny{--} \carAny{--} \carAny{--} \carAny{--} \carNum{17} \carAny{--} \carAny{--} \carNum{20} \newlineCarPlot
 \carPlotLine{}\carAny{--} \carAny{--} \carAny{--} \carAny{--} \carAny{--} \carNum{06} \carAny{--} \carNum{02} \carAny{--} \carAny{--} \carAny{--} \carAny{--} \carAny{--} \carAny{--} \carAny{--} \carAny{--} \carNum{07} \carAny{--} \carAny{--} \carAny{--} \driveRight{} \carNum{01} \carAny{--} \carAny{--} \carAny{--} \carAny{--} \carNum{06} \carAny{--} \carAny{--} \carAny{--} \carAny{--} \carNum{11} \carAny{--} \carAny{--} \carAny{--} \carAny{--} \carAny{--} \carNum{17} \carAny{--} \carAny{--} \carNum{20} \newlineCarPlot
 \carPlotLine{}\carAny{--} \carAny{--} \carAny{--} \carAny{--} \carAny{--} \carAny{--} \carNum{06} \carAny{--} \carNum{02} \carAny{--} \carAny{--} \carAny{--} \carAny{--} \carAny{--} \carAny{--} \carAny{--} \carAny{--} \carNum{07} \carAny{--} \carAny{--} \driveRight{} \carNum{01} \carAny{--} \carAny{--} \carAny{--} \carAny{--} \carNum{06} \carAny{--} \carAny{--} \carAny{--} \carAny{--} \carNum{11} \carAny{--} \carAny{--} \carAny{--} \carAny{--} \carAny{--} \carNum{17} \carAny{--} \carAny{--} \carNum{20} \newlineCarPlot
 \carPlotLine{}\carAny{--} \carAny{--} \carAny{--} \carAny{--} \carAny{--} \carAny{--} \carAny{--} \carNum{06} \carAny{--} \carNum{02} \carAny{--} \carAny{--} \carAny{--} \carAny{--} \carAny{--} \carAny{--} \carAny{--} \carAny{--} \carNum{07} \carAny{--} \driveRight{} \carNum{01} \carAny{--} \carAny{--} \carAny{--} \carAny{--} \carNum{06} \carAny{--} \carAny{--} \carAny{--} \carAny{--} \carNum{11} \carAny{--} \carAny{--} \carAny{--} \carAny{--} \carAny{--} \carNum{17} \carAny{--} \carAny{--} \carNum{20} \newlineCarPlot
 \carPlotLine{}\carAny{--} \carAny{--} \carAny{--} \carAny{--} \carAny{--} \carAny{--} \carAny{--} \carNum{06} \carAny{--} \carNum{02} \carAny{--} \carAny{--} \carAny{--} \carAny{--} \carAny{--} \carAny{--} \carAny{--} \carAny{--} \carNum{07} \carAnyYield{--} \driveLeft{} \carFree{01} \carAny{--} \carAny{--} \carAny{--} \carAny{--} \carNum{06} \carAny{--} \carAny{--} \carAny{--} \carAny{--} \carNum{11} \carAny{--} \carAny{--} \carAny{--} \carAny{--} \carAny{--} \carNum{17} \carAny{--} \carAny{--} \carNum{20} \newlineCarPlot
 \carPlotLine{}\carAny{--} \carAny{--} \carAny{--} \carAny{--} \carAny{--} \carAny{--} \carAny{--} \carNum{06} \carAny{--} \carNum{02} \carAny{--} \carAny{--} \carAny{--} \carAny{--} \carAny{--} \carAny{--} \carAny{--} \carAny{--} \carNum{07} \carAny{--} \driveLeft{} \carAny{--} \carAny{--} \carAny{--} \carAny{--} \carNum{05} \carAny{--} \carAny{--} \carAny{--} \carAny{--} \carNum{10} \carAny{--} \carAny{--} \carAny{--} \carAny{--} \carAny{--} \carNum{16} \carAny{--} \carAny{--} \carNum{19} \carAny{--} \newlineCarPlot